\documentclass{article}
\usepackage{spconf,amsmath,graphicx,hyperref}

\usepackage{booktabs} 
\usepackage{multirow}
\usepackage{amsmath}
\usepackage{amssymb}
\usepackage{pifont}
\usepackage{float}
\usepackage{marvosym}
\title{LPCVAE: A Conditional VAE with Long-Term Dependency and Probabilistic Time-Frequency Fusion for Time Series Anomaly Detection}
\name{Hanchang Cheng$^{\star \dagger}$ \qquad Weimin Mu$^{\star}$ \qquad Fan Liu$^{\star *}$ \qquad Weilin Zhu$^{\star \dagger}$ \qquad Can Ma$^{\star}$ \thanks{*Corresponding author: liufan@iie.ac.cn}}
  \address{$^{\star}$ Institute of Information Engineering, Chinese Academy of Sciences, Beijing, China\\
      $^{\dagger}$ School of Cyber Security, University of Chinese Academy of Sciences, Beijing, China}

\begin{document}
\ninept
\maketitle
\begin{abstract}
Time series anomaly detection(TSAD) is a critical task in signal processing field, ensuring the reliability of complex systems. Reconstruction-based methods dominate in TSAD. Among these methods, VAE-based methods have achieved promising results. Existing VAE-based methods suffer from the limitation of single-window feature and insufficient leveraging of long-term time and frequency information. We propose a \textbf{C}onditional \textbf{V}ariational \textbf{A}uto\textbf{E}ncoder with \textbf{L}ong-term dependency and \textbf{P}robabilistic time-frequency fusion, named \textbf{LPCVAE}. LPCVAE introduces LSTM to capture long-term dependencies beyond windows. It further incorporates a Product-of-Experts (PoE) mechanism for adaptive and distribution-level probabilistic fusion. This design effectively mitigates time-frequency information loss. Extensive experiments on four public datasets demonstrate it outperforms state-of-the-art methods. The results confirm that integrating long-term time and frequency representations with adaptive fusion yields a robust and efficient solution for TSAD.  Our code is publicly available at \url{https://github.com/Cache233/LPCVAE}.
\end{abstract}
\begin{keywords}
Time Series, Anomaly Detection, Conditional Variational AutoEncoder, Time-Frequency Analysis.
\end{keywords}
\section{Introduction}
\label{sec:intro}

Time series anomaly detection is a fundamental technique in signal processing field. It identifies data points that significantly deviate from expected behavioral patterns in time series. The timely detection of anomalies, such as service outages or cyberattacks, is critical for maintaining the stability and security of network systems.

\begin{figure}[t]
    \centering
    \includegraphics[height=0.27\textheight, width=0.8\linewidth]{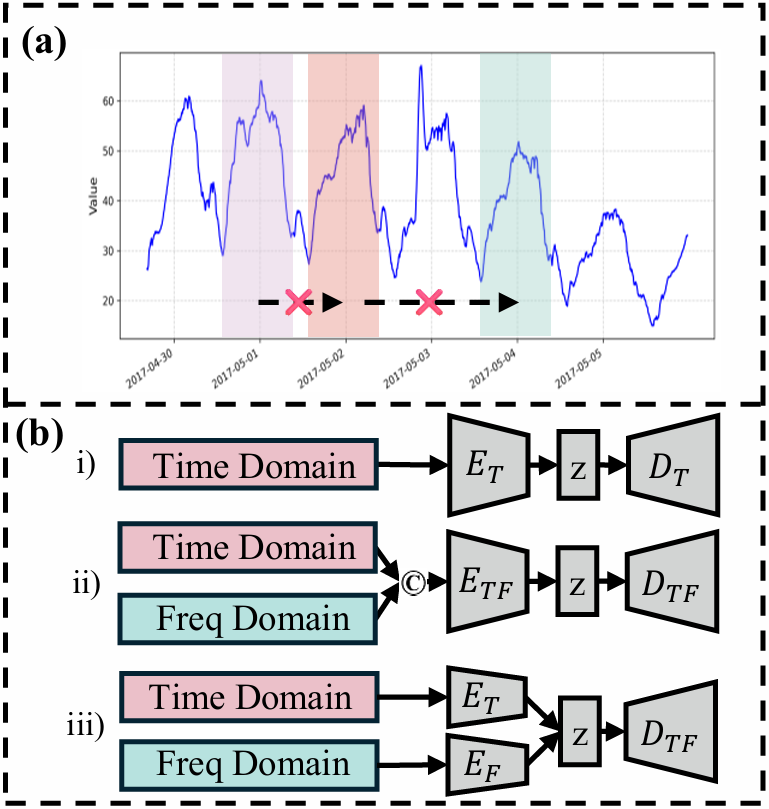}
    \caption{(a) Long-term historical information cannot be propagated across input windows. (b)VAE-based Methods comparison: i) Time-only architecture. ii) Time-Frequency concatenation architecture. iii) Our LPCVAE architecture. Here, © denotes concatenation, and Freq represents the frequency.}
    \label{fig:vae-based Methods comparison}
\end{figure}

Existing methods are primarily categorized into traditional and deep learning-based methods. Traditional methods \cite{weekley2010algorithm, li2003improving, liu2008isolation} include statistical models and conventional machine learning algorithms. These methods are often limited by handcrafted features and sensitivity to hyperparameter settings and noise. Deep learning approaches\cite{ren2019time, su2019robust, yang2023dcdetector} with the ability to learn discriminative representations, are widely applied to anomaly detection. In deep learning methods, reconstruction-based methods emerge as the dominant paradigm. Variational Autoencoders (VAEs)\cite{vae} and Conditional Variational Autoencoders (CVAEs)\cite{cvae} excel in probabilistic latent space modeling. Normal data produces statistically bounded reconstruction errors, whereas anomalies appear as significant deviations. This fundamental alignment with anomaly detection principles position VAE-based Methods at the forefront of current research. However, existing methods still face several challenges: \\
% \textbf{Challenge 1: VAEs Constrained by Single Window dependency.}
% Most existing VAE-based Methods methods typically use a sliding window to divide long time series into smaller segments for training. Each window is modeled independently, ignoring inter-segment dependencies, which are crucial for time series modeling. This limitation reduces the model's ability to capture long-term time dependencies. Therefore, developing efficient frameworks to model these dependencies remains a fundamental research challenge.
\indent\textbf{Challenge 1: VAEs Constrained by Single Window Feature.}
Most VAE-based methods focus on modeling each single window, ignoring inter-window dependency that are critical for time series reconstruction, as illustrated in Fig.~\ref{fig:vae-based Methods comparison}(a). VAE-LSTM\cite{lin2020anomaly} attempts to address this issue by applying recurrent modeling in the latent space. However, it still relies on single-window inputs and captures temporal feature only after compression, limiting the expressive capacity of long-term dependency. Therefore, modeling these dependencies remains a key challenge.
% \begin{table}[htbp]
% \centering
% \caption{VAE-based Methods Method Comparison.}
% \begin{tabular}{c|c|c|c}
% \toprule
% Method & Time & Frequency & Fusion strategy\\
% \midrule
% Donut & \checkmark & \ding{55} & \textbf{--}\\
% VQRAE & \checkmark & \ding{55} & \textbf{--}\\
% FCVAE & \checkmark & \checkmark & concatenation\\
% \textbf{Our Work} & \checkmark & \checkmark & PoE \\
% \bottomrule
% \end{tabular}
% \label{tab:vaes-based Method comparison}
% \end{table}

\begin{figure*}[htbp]
    \centering
    \includegraphics[width=1\linewidth]{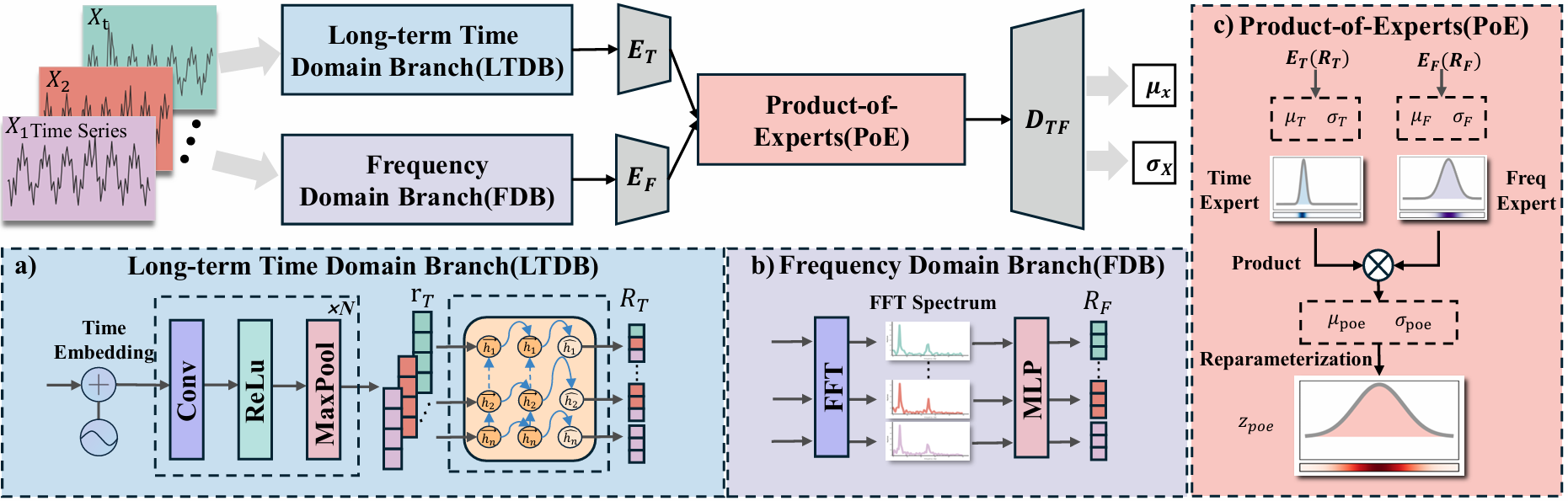}
    \caption{LPCVAE Model Architecture.}
    \label{fig:architecture}
\end{figure*}

\textbf{Challenge 2: Lack of Effective Leveraging of Time and Frequency Information.} Existing VAE-based methods adopt different strategies for handling time and frequency information, as shown in Fig.~\ref{fig:vae-based Methods comparison}(b). These methods face significant limitations in time-frequency integration. Time-only methods (e.g., Donut\cite{xu2018unsupervised}, VQRAE\cite{KieuYGCZSJ22}) ignore frequency-domain features critical for capturing periodic patterns. Dual-domain methods like FCVAE\cite{wang2024revisiting} rely on simple concatenation. FCVAE only partially leverages frequency features and fails to model interactions between time and frequency domains. The incomplete and disjoint exploitation of both domains hinders the model’s ability to fully capture the underlying patterns in time series. As a result, the performance of anomaly detection is constrained. These challenges highlight the need for more advanced fusion strategies to achieve effective feature integration.\\
To address the aforementioned challenges, we propose LPCVAE, a Conditional VAE that captures long-term dependencies and fuses time-frequency representations probabilistically. The main contributions are as follows:
\begin{itemize}
    \item To address the constraint of single-window feature in existing VAEs, we propose a LSTM-based long-term dependency extraction mechanism. This mechanism enhances the modeling scope by propagating historical information across windows within each training batch.
    \item We pioneer the application of the Product-of-Experts (PoE) mechanism to time-frequency feature fusion in time series analysis. This approach establishes a probabilistic joint model of time and frequency domain representations. It achieves more effective and adaptive feature integration.
    \item Evaluation results demonstrate that LPCVAE significantly outperforms state-of-the-art methods on four public datasets, with the largest improvement of 6.3\% on the Yahoo dataset. In addition, ablation studies and analyses on key components further confirm the effectiveness of our method and explain the reasons behind its performance gains.
\end{itemize}

\begin{table*}
\centering
\caption{Contrast Experiments on Public Benchmark Datasets}
\begin{tabular*}{\textwidth}{@{\extracolsep{\fill}} c|ccc|ccc|ccc|ccc}
\toprule
\multirow{2}{*}{Method} & 
\multicolumn{3}{c}{NAB} & 
\multicolumn{3}{c}{Yahoo} & 
\multicolumn{3}{c}{KPI} & 
\multicolumn{3}{c}{WSD} \\
& {P} & {R} & {F1} & 
  {P} & {R} & {F1} & 
  {P} & {R} & {F1} & 
  {P} & {R} & {F1} \\
\midrule
% 插入你的数据行，例如：
% SPOT & 0.992 & 0.713 & 0.829 & 0.572 & 0.328 & 0.417 & 0.966 & 0.221 & 0.360 & 0.947 & 0.315 & 0.472\\
% Donut & 0.933 & 0.937 & 0.935 & 0.381 & 0.150 & 0.215 & 0.378 & 0.569 & 0.454 & 0.263 & 0.195 & 0.224\\
% VQRAE & 0.990 & 0.882 & 0.933 & 0.706 & 0.399 & 0.510 & 0.202 & 0.418 & 0.272 & 0.518 & 0.233 & 0.312\\

% SRCNN & 0.825 & 0.832 & 0.828 & 0.268 & 0.236 & 0.251 & 0.673 & 0.944 & 0.786 & 0.093 & 0.903 & 0.170\\
Anomaly-Transformer & 0.868 & 1.000 & 0.930 & 0.127 & 0.156 & 0.140 & 0.562 & 0.497 & 0.527 & 0.130 & 0.306 & 0.182\\
TFAD & 0.922 & 1.000 & 0.959 & 0.774 & 0.339 & 0.472 & 0.835 & 0.753 & 0.791 & 0.667 & 0.432 & 0.525\\
TFMAE & 0.798 & 1.000 & 0.888 & 0.192 & 0.239 & 0.213 & 0.593 & 0.825 & 0.690 & 0.293 & 0.139 & 0.189\\
MENTO & 0.970 & 0.478 & 0.640 & 0.007 & 0.007 & 0.007 & 0.913 & 0.430 & 0.585 & 0.338 & 0.101 & 0.155\\
TimeMixer & 0.897 & 0.356 & 0.510 & 0.051 & 0.016 & 0.025 & 0.749 & 0.320 & 0.448 & 0.183 & 0.127 & 0.150\\
FCVAE & 0.942 & 1.000 & \underline{0.970} & 0.896 & 0.820 & \underline{0.856} & 0.897 & 0.942 & \underline{0.919} & 0.811 & 0.805 & \underline{0.808}\\
\midrule
LPCVAE & 0.989 & 1.000 & \textbf{0.995} & 0.903 & 0.936 & \textbf{0.919} & 0.943 & 0.941 & \textbf{0.942} & 0.901 & 0.819 & \textbf{0.858} \\
\bottomrule
\end{tabular*}
\label{tab:contrast}
\end{table*}

\section{METHODOLOGY}
\label{sec:format}
\subsection{Problem Formulation}
Given a time series $X = \{x_0, x_1, \ldots, x_n\}$, where $x_n \in \mathbb{R}$, $n \in \mathbb{N}$. The time window is defined as $X_t = \{x_t, x_{t+1}, \ldots, x_{t+w-1}\}$, where $w$ denotes the sliding window size and $t \in [0, n-w+1]$. The objective is to identify a corresponding detection label sequence 
$\mathbf{L}_t = \{l_t, l_{t+1}, \ldots, l_{t+w-1}\}$, 
where each $l_t \in \{0, 1\}$ denotes the label at time $t$, 
with $0$ indicating normal and $1$ indicating anomalous.
\subsection{Network Architecture}
Our framework, as illustrated in Fig.~\ref{fig:architecture}, is composed of three core components: a \textbf{Long-term Time Domain Branch(LTDB)} for extracting sequential dependencies, a \textbf{Frequency Domain Branch(FDB)} for capturing periodic structures, and a \textbf{CVAE Based Product of Experts} to integrate long-term time and frequency representations.
\subsection{Long-term Time Domain Branch(LTDB)}
Our LTDB models both short-term and long-term dependencies in time series. First, we introduce Time Embedding(TME), which combines Position Embedding (PE) and Timestamp Embedding (TE). PE preserves the sequential order within each input window. TE incorporates real-world time signals to capture periodicity and situational information. Together, they enrich temporal representations for subsequent convolutional operations. Then, the encoded sequences are processed by a 1D-Conv with ReLU activation and max-pooling. The extraction of local window-level representations is formulated as follows:
\begin{equation}
r_T(t) = \mathit{MaxPool}\left(\delta\left(W_c * TME(X_t) + b_c\right)\right)
\end{equation}
where $r_T(t)$ denotes the local feature representation of a single window $X_t$, $\text{TME}(x)$ denotes the input’s time embedding, $W_c$ and $b_c$ are convolution parameters, $*$ is the convolution operator, and $\delta(\cdot)$ is the ReLU activation.  

To overcome the limited feature of single window, we introduce LSTM to process the extracted features sequentially across the entire batch. Leveraging its recurrent state, LSTM integrates both local and historical features from preceding windows. It is formulated as follows:
\begin{equation}
R_T(t) = \mathit{LSTM}\left(r_T(t), H(r_T(t-1)), C(r_T(t-1))\right)
\end{equation}
where $R_T(t)$ denotes the global feature representation of window $X_t$, and $H(\cdot)$ and $C(\cdot)$ denote the hidden state and cell state of the LSTM, respectively. 
\subsection{Frequency Domain Branch(FDB)}
The frequency domain branch extracts spectral periodicity and dominant frequency components. These features are hard to capture directly in time domain analysis but play a critical role in anomaly detection. The time series is converted into the frequency domain using the Fast Fourier Transform (FFT) to preserve full spectral information. Noise and anomalies typically appear as long tails in the spectrum. To address this problem, a Multi-Layer Perceptron (MLP) with Dropout is applied to perform nonlinear transformations. It effectively suppresses noise and anomalies while enhancing discriminative features. The frequency-domain feature  is formulated as:  
\begin{equation}
\label{eq:freq}
R_F(t) = \mathit{Dropout}\left( \mathit{MLP} \left( \mathit{FFT}(X_t) \right) \right)
\end{equation}

\subsection{CVAE Based Product of Experts}
The time-domain feature $R_T(t)$ and frequency-domain feature $R_F(t)$ are encoded by the time encoder $E_T$ and frequency encoder $E_F$. The encoders output corresponding latent distribution parameters $(\mu_T, \sigma_T)$ and $(\mu_F, \sigma_F)$, respectively.  

Conventional studies often adopt decision-level fusion, such as fixed threshold\cite{tfad} or anomaly scores summation\cite{nam2024breaking}. These methods typically incur information loss and lack adaptability. To address these issues, we employ the Product of Experts (PoE)\cite{hinton2002training} for distribution-level fusion. It models time and frequency domains while adaptively weighting their contributions.

Formally, the time domain branch and frequency domain branch are treated as independent probabilistic experts. Under the PoE framework, their posteriors are fused by normalizing the product of the individual distributions, which yields updated mean and variance. The formulation is presented below:
\begin{equation}
\frac{1}{\sigma_{\mathit{poe}}^2} = \frac{1}{\sigma_T^2} + \frac{1}{\sigma_F^2}
\end{equation}
\begin{equation}
\mu_{\mathit{poe}} = \sigma_{\mathit{poe}}^2 \cdot \left( \frac{\mu_T}{\sigma_T^2} + \frac{\mu_F}{\sigma_F^2} \right)
\end{equation}

The latent representation $z_{poe}$ is sampled via the reparameterization trick, which is given by:
\begin{equation}
z_{\mathit{poe}} = \mu_{\mathit{poe}} + \sigma_{\mathit{poe}} \cdot \epsilon, \quad \epsilon \sim \mathcal{N}(0, I)
\end{equation}
\indent Finally, the decoder reconstructs the time series $\widehat{X}_t$ from $z_{\mathit{poe}}$, which is given by:
\begin{equation}
\mu_x, \sigma_x = D_{\mathrm{TF}}(z_{\mathit{poe}})
\end{equation}
\begin{equation}
\widehat{X}_{t} = \mu_{\text{x}} + \sigma_{\text{x}} \cdot \epsilon, \quad \epsilon \sim \mathcal{N}(0, I)
\end{equation}
\subsection{Training and Testing}
In the training phase, the objective minimizes both reconstruction loss and the KL divergence between the latent distribution and a standard Gaussian, which is defined as follows:
\begin{equation}
\label{eq:total_loss}
\mathcal{L} = \underbrace{\text{Reconstruction Loss}}_{\text{L1}} 
+ \underbrace{\text{KL Divergence}}_{\text{L2}}
\end{equation}
\indent FCVAE proposed the condition modified ELBO (CM-ELBO) to weaken the impact of abnormal and missing data in the window on the reconstruction. We adapte it accordingly to better align with our PoE mechanism, and is given by:
\begin{equation}
\label{eq:Reconstruction Loss}
\text{L1} = \mathbb{E}_{q_{\phi}(z_{t} \mid X_t, c_t)} 
\Bigg[ \sum_{i=0}^{w-1} \alpha_{t+i} \log p_{\theta}(x_{t+i} \mid z_t, c_t) \Bigg]
\end{equation}
\begin{equation}
\label{eq:KL Divergence}
\text{L2} = \mathbb{E}_{q_{\phi}(z_{t} \mid X_t, c_t)} 
\Big[ \beta_t \log p_{\theta}(z_t \mid X_t) - \log q_{\phi}(z_t \mid X_t, c_t) \Big]
\end{equation}
where $z_t$ denotes the latent representation of $X_t$, and $c_t$ denotes the condition information incorporating long-term temporal and frequency features. $q_{\phi}(\cdot)$ and $p_{\theta}(\cdot)$ are the variational encoder and decoder parameterized by $\phi$ and $\theta$, respectively. $\alpha_{t+i}=1$ if $x_{t+i}$ is normal, and $\alpha_{t+i}=0$ otherwise, and $\beta_t = \left(\sum_{i=0}^{w-1}\alpha_{t+i}\right) / w$ denotes the proportion of valid observations in the window.\\
\indent During testing, we quantify reconstruction difficulty by using reconstruction probabilities as anomaly score. The anomaly score is defined as:
\begin{equation} \label{eq:Anomaly Score}
\text{AnomalyScore}(X_t) 
= -\mathbb{E}_{q_{\phi}(z_t \mid X_t, c_t)} 
\Big[ \log p_{\theta}(X_t \mid z_t, c_t) \Big]
\end{equation}
\section{EXPERIMENT}
\label{sec:pagestyle}
\subsection{Datasets}
To evaluate our method, we use the following four datasets in experiments. The key information about the datasets is summarized in the Table~\ref{tab:datasets}.
\begin{table}[H]
\centering
\caption{Summary of Datasets.}
\begin{tabular}{c|c|c|c}
\toprule
Dataset & Points & Anomaly Points & Anomaly Rate \\
\midrule
NAB\cite{nab} & 158631 & 15684 & 9.89\% \\
Yahoo\cite{yahoo} & 572966 & 3915 & 0.68\% \\
KPI\cite{aiops} & 5922913 & 134114 & 2.26\% \\
WSD\cite{wsd} & 7529176 & 121279 & 1.61\% \\
\bottomrule
\end{tabular}
\label{tab:datasets}
\end{table}

\subsection{Baselines} We conduct extensive comparisons between LPCVAE and six representative baseline methods. These baselines include: Transformer-based model \textbf{Anomaly-Transformer}\cite{anomaly-transformer}; time–frequency models \textbf{TFAD}\cite{tfad} and \textbf{TFMAE}\cite{fang2024temporal}; the memory-augmented model \textbf{MEMTO}\cite{song2023memto}; the multi-scale analysis model \textbf{TimeMixer}\cite{wang2023timemixer}; as well as VAE-based model \textbf{FCVAE}\cite{wang2024revisiting}.
\subsection{Implementation} We use the Adam optimizer with an initial learning rate of 0.0001 and a batch size of 512. To evaluate anomaly detection performance, we adopt three widely used metrics: precision (P), recall (R), and F1 Score (F1). Following common practice in the literature\cite{xu2018unsupervised}, \cite{wang2024revisiting}, \cite{anomaly-transformer}, the point-adjustment strategy is applied. 
% Under this strategy, a continuous anomaly segment is considered correctly detected if at least one observation within the segment is identified. 
All experiments are conducted on a single NVIDIA GeForce RTX 3090 GPU.
\subsection{Performance Comparison}
Table~\ref{tab:contrast} presents the comparative results of LPCVAE against existing methods across multiple datasets: (1) LPCVAE consistently achieves the best performance under different scenarios, with F1 Score of 0.995 on the NAB dataset remarkably. (2) Compared with state-of-the-art VAE-based method FCVAE, LPCVAE improves F1 Score on the Yahoo dataset by 6.3\%. These results demonstrate the effectiveness and superiority of LPCVAE in the task of time series anomaly detection.
\subsection{Ablation Study}
\begin{figure}[H]
    \centering
    \includegraphics[width=1\linewidth]{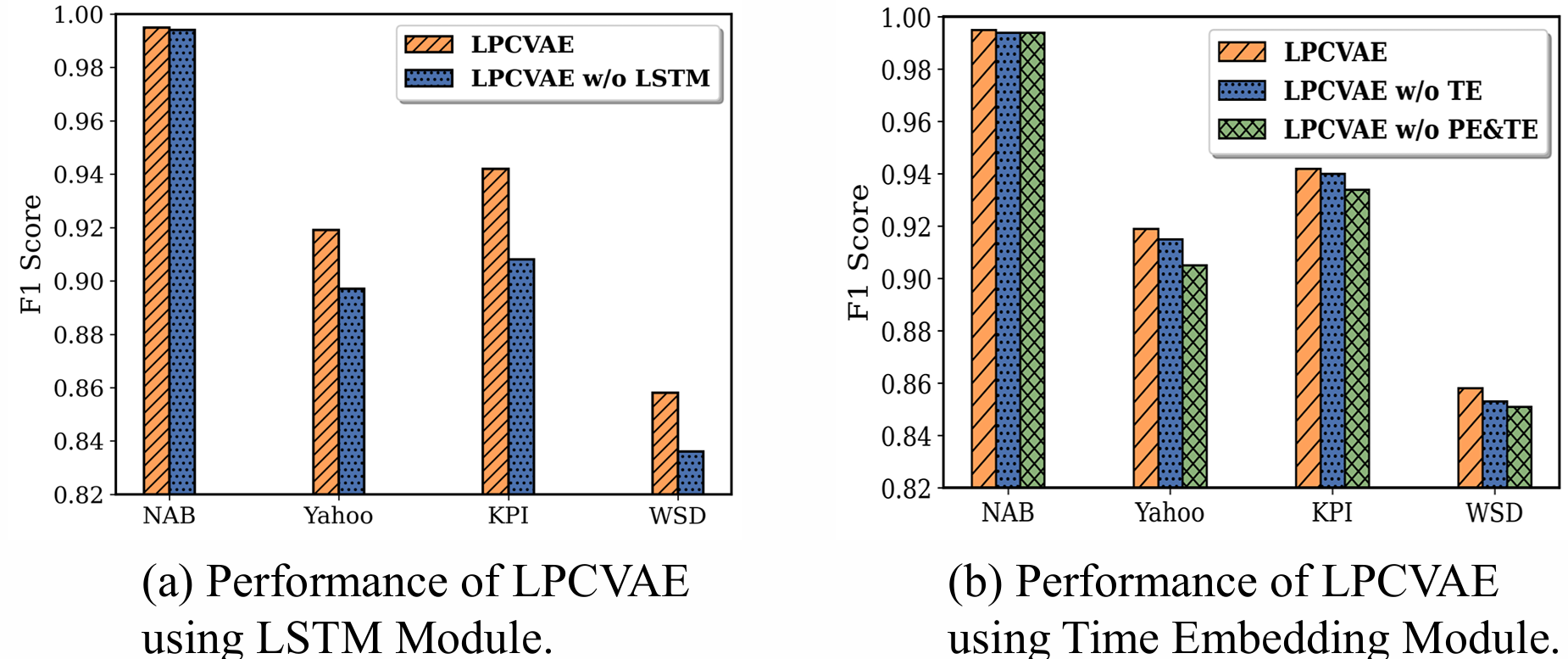}
    \caption{F1 Score of different settings."w/o" denotes "without"; e.g., "w/o LSTM" means LPCVAE without LSTM module. }
    \label{fig:ablation}
\end{figure}

\begin{table}[H]
\centering
\caption{F1 Score of different ablation settings.}
\begin{tabular}{ccc|c|c|c|c}
\toprule
\textbf{LTDB} & \textbf{FDB} & \textbf{PoE} & \textbf{NAB} & \textbf{Yahoo} & \textbf{KPI} & \textbf{WSD} \\
\midrule
\checkmark & \ding{55} & \textbf{--} & 0.993 & 0.885 & 0.884 & 0.796 \\
\ding{55} & \checkmark & \textbf{--}  & 0.993 & 0.881 & 0.901 & 0.813 \\
\checkmark & \checkmark & \ding{55} & 0.994 & 0.901 & 0.924 & 0.834 \\
\checkmark & \checkmark & \checkmark & \textbf{0.995} & \textbf{0.919} & \textbf{0.942} & \textbf{0.858} \\
\bottomrule
\end{tabular}
\label{tab:ablation}
\end{table}
\textbf{Time-Frequency Joint Architecture.} As shown in Table~\ref{tab:ablation}, our LPCVAE achieves an average improvement of 3.9\% in F1 Score over the LTDB-only model and 3.15\% over the FDB-only model across four datasets. The model leverages both long-term time and frequency information, consistently outperforming the single-perspective variants. \\
\textbf{PoE Mechanism.} Compared with simple concatenation, PoE achieves the best results, yielding an average improvement of 1.5\% across four datasets. This demonstrates the advantage of distribution-level fusion in preserving complementary information and reducing information loss.\\
\textbf{LSTM Module.} As shown in Fig.~\ref{fig:ablation}(a), our method achieves highest gain of 3.4\% on KPI by introducing LSTM. This highlights its effectiveness in modeling long-term dependencies.\\
\textbf{TimeEmbedding.} As shown in Fig.~\ref{fig:ablation}(b), LPCVAE with Time Embedding module achieves the best performance across all datasets. This indicates the joint use of positional embedding (PE) and time embedding (TE) improving model's performance.
\subsection{Hyper-Parameter Sensitivity}
\begin{figure}[H]
    \centering
    \includegraphics[width=0.98\linewidth]{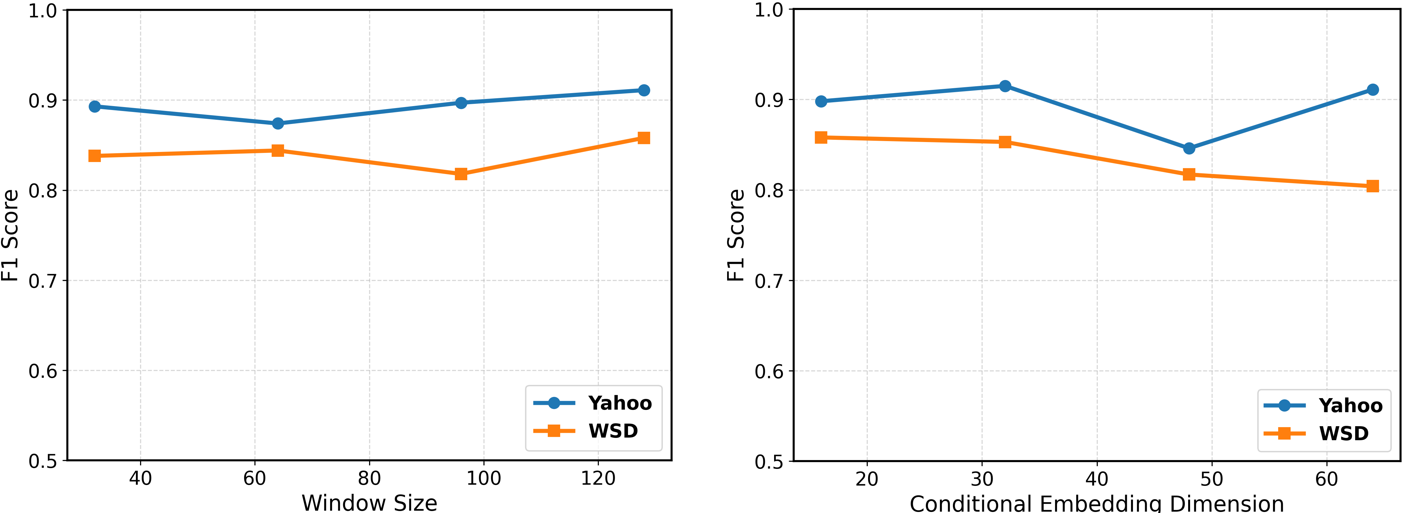}
    \caption{F1 Score of different Hyper-Parameters.}
    \label{fig: ParameterSensitivity}
\end{figure}

We evaluate the sensitivity of model parameters on the Yahoo and WSD datasets, focusing on two aspects: window size and conditional embedding dimension. As shown in Fig.~\ref{fig: ParameterSensitivity}, LPCVAE remains relatively stable and performs consistently well under different parameter settings. For the Yahoo dataset with complex periodicity, a larger conditional embedding dimension improves performance.
\subsection{Computation Efficiency}
\begin{figure}[H]
    \centering
    \includegraphics[width=0.7\linewidth]{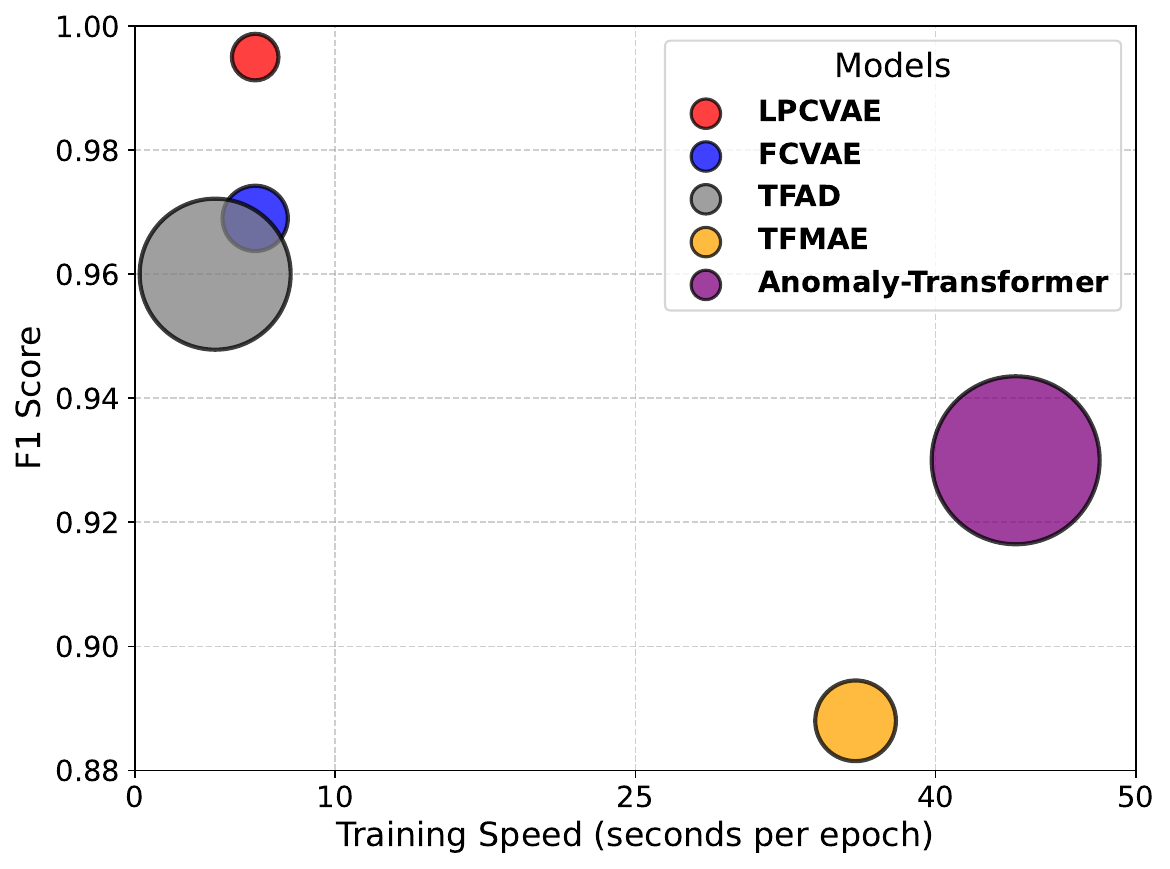}
    \caption{Computation Efficiency Comparison on NAB Dataset.}
    \label{fig:ComputationEfficiency}
\end{figure}

% 为了综合评估LPCVAE的性能和计算效率，我们对FCVAE、TFMAE、Anomaly-transformer、TimeMixer、TFAD多种前沿基线方法，在F1分数、训练速度和GPU内存占用（圆的面积越大，内存占用越高）方面进行了对比。如图\ref{fig:ComputationEfficiency}所示，LPCVAE取得了最高的F1分数，拥有最低的GPU内存占用。同时在训练速度上排名第二，仅次于内存占用更高的TFAD。值得注意的是，LPCVAE的训练速度与目前基于VAEs最好的模型（翻译成sota即可）的FCVAE持平但性能更好，GPU内存占用更低。总体来说，LPCVAE在速度与性能之间实现了更优的权衡，同时具备出色的内存使用效率。
To comprehensively evaluate the performance and computational efficiency of LPCVAE, we compare it with several representative baselines, including FCVAE, TFMAE, Anomaly-Transformer, TimeMixer, and TFAD. The comparison considers F1 Score, training speed, and GPU memory usage, where larger circle areas indicate higher memory consumption. As shown in Fig.~\ref{fig:ComputationEfficiency}, LPCVAE achieves the highest F1 Score while requiring the least GPU memory. In terms of training speed, it ranks second, only behind TFAD which demands significantly more memory. Notably, LPCVAE matches the training speed of the best VAE-based model FCVAE, while delivering superior accuracy with lower memory usage. This demonstrates its favorable balance between speed and performance.
\section{CONCLUSION}
\label{sec:typestyle}
% In this paper, we investigate the limitations of existing VAE-based approaches for time series anomaly detection. We propose LPCVAE, a model that effectively captures and fuses both time and frequency domain information. By leveraging the PoE mechanism, LPCVAE adaptively fuses multi-domain features. The LSTM module addresses the limitation of window-independent learning by modeling long-term dependencies. Extensive experiments on four real-world datasets demonstrate that LPCVAE achieves superior detection accuracy. It provides a valuable solution for time series anomaly detection (TSAD).

In this paper, we propose a novel model \textbf{LPCVAE} for time series anomaly detection, which departs from the architecture of conventional VAE-based approaches. By introducing LSTM to model dependencies across windows, we overcome the inherent limitations of single-window feature. Additionally, we adopt the Product-of-Experts (PoE) mechanism to adaptively fuse time and frequency domain representations at the distribution level. This alleviates the issue of insufficient leveraging of time–frequency information. Extensive experiments on four real-world datasets demonstrate that LPCVAE achieves superior detection performance. It provides a valuable solution for time series anomaly detection.

\bibliographystyle{IEEEbib}
\bibliography{strings,refs}

\end{document}